\documentclass[review]{elsarticle}

\usepackage{hyperref}
\usepackage[hyphenbreaks]{breakurl}

\usepackage{lineno}
\modulolinenumbers[5]
\usepackage{graphicx,subcaption}
\usepackage{array}
\usepackage{tabularx} 
\usepackage[numbers]{natbib}
\usepackage{adjustbox}
\usepackage{amsmath,bm}
\usepackage{amssymb}
\usepackage{siunitx}
\usepackage{textcomp}
\usepackage{mathtools}
\usepackage{tabularx,multirow,makecell}
\usepackage[ruled,vlined]{algorithm2e}
\usepackage{setspace}
\usepackage{xcolor}
\usepackage{color}\usepackage{color}

\usepackage{soul}

\usepackage[flushleft]{threeparttable}

\journal{Expert Systems with Applications}
\biboptions{authoryear}

\usepackage{setspace}
\usepackage{xcolor}
\usepackage{color}
\usepackage{soul}

\def\XS{\xspace}
\def\figureabvr{Figure\XS} 
\def\tableabvr{Table\XS}
\def\eg{\textit{e.g.,}\XS}
\def\etal{\textit{et al.}\XS}
\def\etc{\textit{etc.}\xspace}
\def\ie{\textit{i.e.,}\xspace}

\def\figuredraft{false}

\def\rev#1{{#1}}

\journal{Expert Systems with Applications}

\begin{document}
\begin{frontmatter}

\title{Self-Supervised Endoscopic Image Key-Points Matching}

\author[crns]{Manel Farhat}
\ead{farhat.manel@hotmail.fr}

\author[crns,smart]{Houda Chaabouni-Chouayakh}
\ead{houdachaabouni@gmail.com}

\author[crns,smart]{Achraf Ben-Hamadou\corref{mycorrespondingauthor}}
\ead{achraf.benhamadou@crns.rnrt.tn}

\address[crns]{Centre de Recherche en Numérique de Sfax, Technopôle de Sfax, 3021 Sfax, Tunisia}

\address[smart]{Laboratory of Signals, systems, artificial Intelligence and networks, Technopôle de Sfax, 3021 Sfax, Tunisia}
\cortext[mycorrespondingauthor]{Corresponding author}

\begin{abstract}

Feature matching and finding correspondences between endoscopic images is a key step in many clinical applications such as patient follow-up and generation of panoramic  image from clinical sequences for fast anomalies localization. Nonetheless, due to the high texture variability present in endoscopic images, the development of robust and accurate feature matching becomes a challenging task. Recently, deep learning techniques  which deliver learned features extracted via convolutional neural networks (CNNs) have gained traction in a wide range of computer vision tasks. However, they all follow a supervised learning scheme where a large amount of annotated data is required to reach good performances, which is generally not always available for medical data databases. To overcome this limitation related to labeled data scarcity, the self-supervised learning paradigm has recently shown great success in a number of applications. This paper proposes a novel self-supervised approach for endoscopic image matching based on deep learning techniques. When compared to \rev{standard hand-crafted} local feature descriptors, our method outperformed them in terms of precision and recall. \rev{Furthermore, our self-supervised descriptor provides a competitive performance in comparison to a selection of state-of-the-art deep learning based supervised methods in terms of precision and matching score.}

\end{abstract}


\end{frontmatter}

\doublespacing

\section{Introduction}

Bladder cancer is the sixth most common cancer and the ninth cancer causing mortality for men, and it is the fifteenth cancer causing mortality for both sexes worldwide \citep{Sung20}. Endoscopy remains the gold standard visual examination for detecting bladder cancer in its early stages. It consists in inserting an endoscope (A lighted, tubular instrument) through the urethra into the bladder. The doctor explores the bladder's inner walls by navigating the endoscope to scan the organ for lesions. Despite this being easy to perform, the endoscopic examination has several limitations. The main one being the very limited field of view, in addition to the reduced maneuverability inside the bladder. Because bladder malignancy (\eg lesions) is multi-focal and typically spreads over larger areas than the endoscope field of view, lesion partitions are divided across multiple video frames. As a result, this makes it time consuming and difficult for the doctor to locate and determine the spatial distribution of the lesions. Furthermore, finding an image of interest for diagnosis within a few minutes of the sequence is also a tough task. An intuitive solution to these limitations is to provide the doctor with a panoramic image of the clinical sequence instead of the image sequence. In this case, the extraction of discriminative local features for accurate image matching is a fundamental step in the construction of a panoramic image.

Despite extensive research into panoramic image generation for endoscopy, the development of a robust and accurate feature matching between endoscopic images is particularly difficult due to the specificity of such images. In fact, the texture of the inner surfaces of the bladder varies greatly between patients. Endoscopic images are also typically reddish with a weak texture. This is in addition to the scarcity of data due to ethical concerns and the personal data protection regulations.

There are three basic approaches to endoscopic image matching that are currently being used in the literature: motion-based methods, global image matching methods, and local image matching methods. For motion based methods, we can cite \citep{Hernandez10,6738266,ALI2016425,CHU2020105370,zenteno} who used optical flow based approaches to find key-point correspondences between bladder images. Despite their efficiency in registering consecutive image frames, motion-based matching methods are still sensitive to challenging situations such as lighting changes, weak textures, and large view-point changes. As a result of this, these methods are limited to matching consecutive image frames and cannot be used for images separated in time, such in case of patient follow-up or the closing loop in a same endoscopic video.

Global image matching strategies ensure the global coherence of the obtained panoramic images and penalize discontinuities by using the entire image content (\eg contours, color, graph, \etc) and specific smoothness constraints. For example, Miranda-Luna \etal \citep{MirandaLuna08} proposed a mosaicing algorithm for endoscopic images based on the maximization of the mutual information using the entire image frames. Weibel \etal \citep{WEIBEL20124138} used instead graph-cuts to minimize a global energy function computed on the entire image pixels. Global matching methods are memory-intensive and time-consuming since the energy function is usually computed over  the entire image pixels. Furthermore, they are better suited to consecutive image matching with minor changes as they are sensitive to the initial geometric transformation between the input images. It is also worth noting that the strong assumption on the planarity of organ surfaces is not usually verified.

Local image matching methods, on the other hand, find the correspondence between image key-points by extracting discriminative features from local image data. These methods are computationally efficient, making them suitable for real-time applications. They usually begin by detecting image key-points and then compute a feature vector for each detected key-point using descriptor techniques such as SIFT \citep{Lowe2004} and SURF\citep{Bayaetal2008}. 

Deep learning techniques have gained traction in the field of local feature description and matching in the recent years. In the domain of endoscopy, to the best of our knowledge, no previous work has been proposed to build discriminative local feature descriptors for endoscopic images using deep learning techniques. Nevertheless, we can find deep learning studies focusing on other purposes like endoscopic images denoising \citep{8918994}, polyp classification \citep{Kim21}, and bleeding zone semantic segmentation \citep{8451300}. They are all based on supervised learning schemes that rely heavily on the availability of annotated data.

Unsupervised learning, in contrast to supervised learning, does not require a labeled dataset, which increases its popularity, particularly in the medical imaging field \citep{Li20,Chen20}. 
Recently, the research community has become increasingly interested in self-supervised learning, a new subset of unsupervised methods. Basically, it consists in training neural networks with automatically generated labels known as pseudo labels \citep{9086055}, and without any manual annotation. Self-supervised learning paradigm contributes to overcome the barrier of labeled data scarcity by leveraging the availability of large amounts of unlabeled data.

In this article, we deal with local image matching methods to find correspondences between image key-points for the purpose of generating panoramic images in endoscopy. In particular, we propose the first self-supervised approach for endoscopic image key-points matching based on deep learning techniques. We designed a convolutional neural network (CNN) as a local feature descriptor to transform patches extracted around key-points into a discriminative embedding space for an effective key-points matching. Our main contribution is the design of a training procedure for the proposed CNN model without need of any labeled data. Indeed the training requires only raw video frames. \rev{The source code related to this study is released on  https://github.com/abenhamadou/Self-Supervised-Endoscopic-Image-Key-Points-Matching.git}

The remainder of this paper is organized as follows: First, an overview on existing local handcrafted and learning-based feature descriptors as well as their applications in the medical field is provided in section \ref{sec:related_works}. Then, section \ref{sec:propsoed_approach} presents the proposed self-supervised endoscopic image key-points  matching approach. After that, experiments and results are presented and discussed in section \ref{sec:expermients_results}. Finally, section \ref{sec:conclusions} gives some conclusions and perspectives.

\section{Related works}
\label{sec:related_works}
In this section, we provide an overview on state-of-the-art local feature descriptors, focusing on the trendy shift from handcrafted to deep learning based feature descriptors. Furthermore, we discuss the use of local feature descriptors in the medical field, emphasizing the labeled data scarcity issue.
\subsection{Handcrafted Local Feature Descriptors}

According to \citep{Maetal2021}, handcrafted local feature descriptors can be classified into floating and binary descriptors based on the discriminative vector type. One of the well-known floating descriptors is SIFT \citep{Lowe2004}, originally calculated based on image gradients.
Inspired by SIFT, \citep{Bayaetal2008} proposed later the SURF descriptor which is much faster than SIFT. SURF used  integral images and Haar wavelet responses in a circular neighborhood around the SURF key-points, which allows to reduce computational costs and speed up the descriptor extraction. \citep{Alcantarillaetal2012} proposed KAZE algorithm whose detection pipeline is similar to SURF. However, it used the MU–SURF method \citep{Agrawaletal2008} to create a non-linear scale space after applying a non-linear diffusion filter. Despite the efficiency of floating descriptors, they are still not suitable for real time applications because of their important computational time. Hence, the rise of binary descriptors. The main benefits of these descriptors are their ease of implementation, simplicity and efficiency. The key idea behind is to generate a binary feature vector by comparing and encoding the intensity of each pixel relatively to its neighbors \citep{Pietikainenetal2011}. Among these methods, we cite BRIEF \citep{Calonderetal2010}, ORB \citep{Rubleeetal2011}, an extended version of BRIEF with rotation invariance and  AKAZE \citep{Alcantarillaetal2013}, an accelerated method of KAZE algorithm.  Another example is  BRISK \citep{6126542}, which exploits a concentric circles  pattern and presents an optimization of BRIEF and ORB.

When it comes to the medical context, handcrafted local feature descriptors are widely used for  image matching \citep{Saha16,2087864}. In \citep{8036841}, the authors proposed a retinal image registration method which combined local feature descriptors with vascular bifurcation. They tested SIFT, SURF and Harris-PIIFD \citep{5416285} as feature descriptors and showed that the combination of SIFT  with vascular bifurcations outperforms other combinations. For retinal image mosaicing, Jalili \etal \citep{Jalili20} used SIFT descriptor to extract local features and then selected optimal features based on a Voronoi diagram. In fluorescence endoscopy, Behrens \etal \citep{Behrens11,Behrens09} proposed a real-time bladder mosaicing method based on SURF feature descriptor. Du \etal \citep{2087864}  designed a SIFT-based zone matching approach for endoscopic images. This approach improves matching results especially with regards to computing time. Also, the work of \citep{fnbot.2022.840594} proposed an improved feature point pair purification algorithm for endoscopic image matching based on the SIFT descriptor. In the same vein, Zhang \etal \citep{ZHANG2022103261} proposed to improve the standard ORB-oriented algorithm using the Gaussian Pyramid method  for endoscopic image mosaicing purposes. It should  be noted that the design of a suitable local feature descriptor plays a critical role in this family of methods.
\subsection{Deep Learning based Local Feature Descriptors}
Among the first successful learning-based image matching approaches, we can cite MatchNet \citep{Han2015MatchNetUF}. It is presented as the combination of two networks: the first one is the feature extracting network, inspired by Siamese network, and the second one is the learned metric network composed of 3 fully connected layers. DeepDesc \citep{7410379} proposed a Siamese network with L2 distance trained by selecting only hardest pairs samples to match in order to increase the descriptor performance. 
Similarly, in \citep{8100132}, the authors proposed a network named L2-Net with seven convolutional layers that outperformed traditional descriptors, including SIFT. GeoDesc \citep{Luo2018GeoDescLL} proposed to learn local descriptors by  including geometry constraints from multi-view reconstructions, achieving thus significant improvements in terms of loss computation, data sampling and data generation during the learning process.
In designing SOSNet, Tian \etal \citep{sosnet2019cvpr} used second-order similarity  into the learning of local descriptors, achieving state-of-the-art performance on several standard benchmarks for different tasks. Later, the same authors \citep{hynet2020}  proposed HyNet, a modified version of L2-Net where all batch normalization layers are replaced by  the off-the-shelf Filter Response Normalization (FRN) layers, which outperforms previous architectures on standard benchmarks.

\rev{Recently, a dense feature descriptor namely DGD-net is designed in \citep{DGDnet2021} where the  training is guided by the
reliability of the descriptor in matching. DGD-net proposes a backtracking method to enhance the localization accuracy.
The authors of  \citep{patch2pix2021} designed Patch2Pix  in a detect-to-refine manner,  which begins with establishing correspondences between patches and then regresses pixel
correspondences according to matched patches using a local search.
A novel local descriptors extraction method named CNDesc is introduced in \citep{CNDesc9761761} where cross normalization technology is used as an alternative to the common L2 normalization. An efficient feature
reuse backbone is designed providing the network with a strong descriptive ability and an image-level distribution consistent loss  is used for regularization
 to enhance the robustness and stability of local descriptors. }

Recently, in \citep{Wiles2021CVPR}, authors proposed a new image matching approach using a co-attention module to condition learned descriptors on both images and a distinctiveness score computed to select the best matches at test time, leading thus to an improved correspondence between image pairs under challenging conditions. 
Sarlin \etal \citep{sarlin2020superglue} designed SuperGlue, an attention-based graph neural networks for local feature matching based on  transformer \citep{NIPS2017}. Inspired by SuperGlue, LoFTR \citep{sun2021loftr} used self and cross attention layers to extract local feature descriptors and match images. It used a linear transformer to reduce the computational complexity.

Motivated by this success of deep learning techniques in various computer vision tasks, the medical imaging community has investigated the transition from handcrafted based systems to learning based systems \citep{7783289, Lui19}. However, this transition has been gradual over the past few years. 
In \citep{7783289}, Khan \etal presented a comparison between handcrafted and CNN features in medical image modality classification based on local image features. They showed that handcrafted features outperform CNN features. This finding was explained by the data intensive nature of the used architecture which make it not enough discriminant when using a limited amount of training data. Luo \etal \citep{8599448} have shown also that CNNs require a relatively large amount of training data to achieve high accuracy. However, the scarcity of labeled data especially in the medical field is a major bottleneck for training such models.

Transfer learning from natural images has been widely used in medical imaging as one of the potential options to mitigate this inherent data issue \citep{Liu20, 7950523, ab869f}. Tajbakhsh \etal \citep{7426826} proposed to fine-tune adequately a pre-trained network for colonoscopy frame classification. Promising results were obtained despite the difference between medical images and ImageNet dataset \citep{5206848} on which the network has been trained. Several studies show also that transfer learning can improve performances in medical applications \citep{10134523,8037242, morid2021scoping}.

Most recently, self-supervised learning has gained popularity as it is a solution to deal with the problem of scarce labeled data. This method has shown a great success in several applications \citep{misra2020self, jing2020self, goyal2019scaling}, but less attention in medical image analysis \citep{spitzer2018improving,bai2019self,zhuang2019self, 9086055}. Azizi \etal \citep{Azizi_2021_ICCV} showed that self-supervised strategy, with unlabeled medical images, significantly outperforms transfer learning strategy.
\section{Proposed approach}
\label{sec:propsoed_approach}
In this section, we first outline the general principle of our proposed matching method for determining the correspondence between endoscopic image key-points. Then we go over our model architecture in detail, as well as the training steps.

\subsection{Image matching approach}
\begin{figure}[h!]
\centering
\includegraphics[draft=\figuredraft,width=\textwidth]{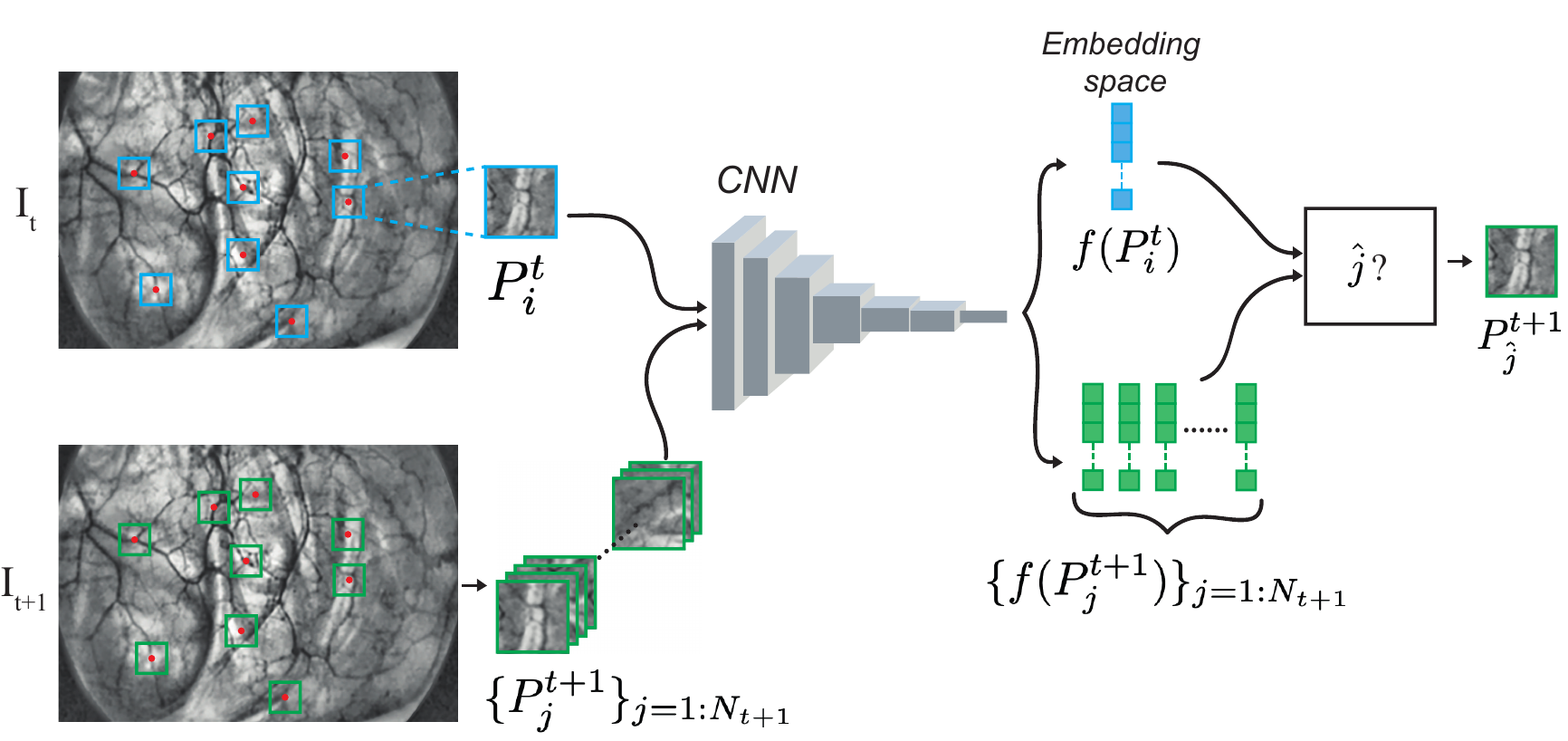}
\caption{Overview of the matching process.}
\label{Fig:Matching}
\end{figure}

The general principal of our image matching approach is depicted in \figureabvr \ref{Fig:Matching}. Let $I_{t}$ and $I_{t+1}$ be two consecutive endoscopic images to be matched. Image key-point detector is applied to the input two images to extract $N_{t}$ and $N_{t+1}$ key-points located in $I_{t}$ and $I_{t+1}$ respectively. We then crop around each key-point a squared patch of $128\times128$ pixels yielding in two sets of patches $\{P^{t}_{i}\}_{i=1:N_{t}}$ and $\{P^{t+1}_{j}\}_{j=1:N_{t+1}}$. The aim of the remaining processing is to find the correspondence between these two sets of patches. To do this, we use a CNN to transform each patch into a more discriminative representation space, and then solve the matching problem by minimizing the Euclidean distance between patches in that embedding space following equation \ref{eq:1}. In \ref{eq:1}, $f(.)$ represents the CNN transformation of a given patch, $\hat{j} \in [1:N_{t+1}]$ is the optimal index matching the $i$-th patch from  $I_{t}$, and $\left\Vert.\right\Vert_2$ stands for the Euclidean distance.

\begin{equation}
\hat{j}=\underset{j} {\mathrm{argmin}} \left\Vert f(P^{t+1}_{j})- f(P^{t}_{i})\right\Vert_2
\label{eq:1}
\end{equation}

\subsection{CNN model architecture}
\begin{figure}[h!]
\centering
\includegraphics[draft=\figuredraft,scale=0.5]{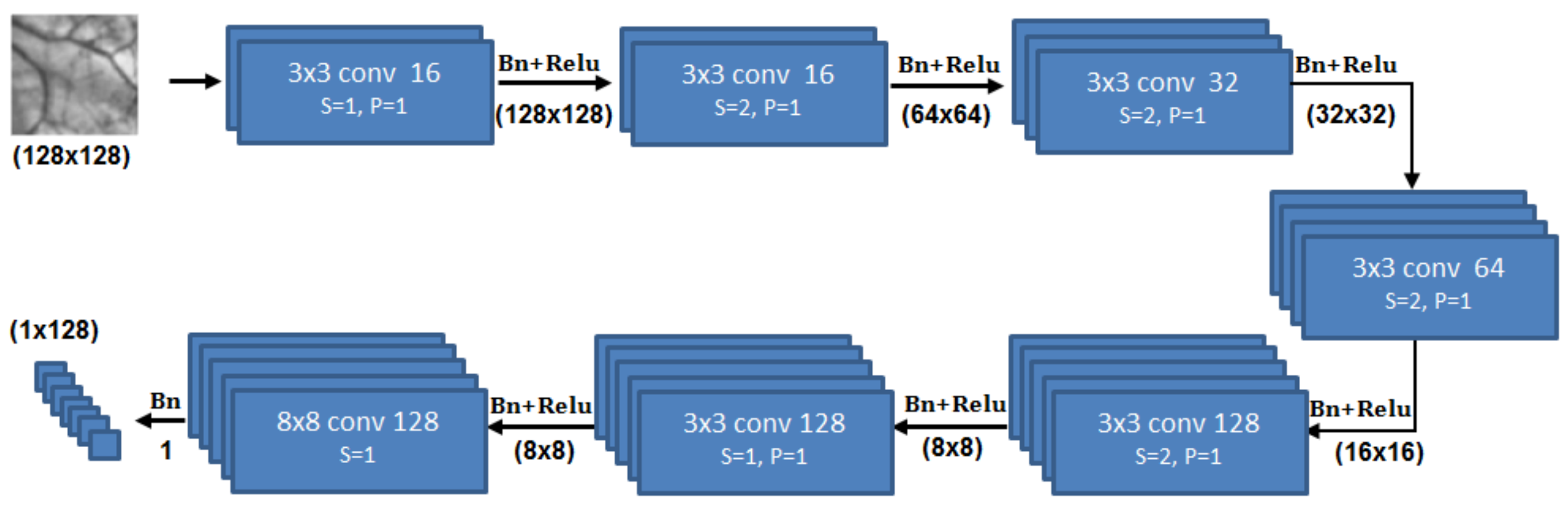}
\caption{Network architecture.}
\label{model}
\end{figure}

The architecture of our network is inspired by L2-Net \citep{8100132}. It consists of seven convolutional layers with 128-D feature vector output. As shown in \figureabvr \ref{model}, the first six convolution layers have small kernel size $(3\times3)$ and are followed by batch normalization and ReLU. The last layer has a kernel size of $(8\times8)$ and is followed only by batch normalization. Batch normalization is considered only for CNN training phase. The number of filters by convolution layer is respectively $\left\{16, 16, 32, 64, 128, 128, 128\right\}$. The padding is set to 1 for all layers (except in the last layer). We used convolution stride equals to 2 instead of pooling layers.

\subsection{CNN model training}
In our approach, the CNN model training phase is critical because it must produce a model with high discriminative capability without using labeled data. A typical supervised-learning scheme would rely entirely on key-point matching ground truth in addition to the input video frames for training. Instead, we designed a self-supervised training approach based on a triplet loss architecture that only requires raw endoscopic video frames. Indeed, triplet loss, introduced in \citep{Schroff_2015} as FaceNet model, has been successfully used in several tasks \citep{Grati20,harvill2019retrieving,kumar2021end}. 
\begin{figure}[h!]
\centering
\includegraphics[draft=\figuredraft,width = 1\textwidth]{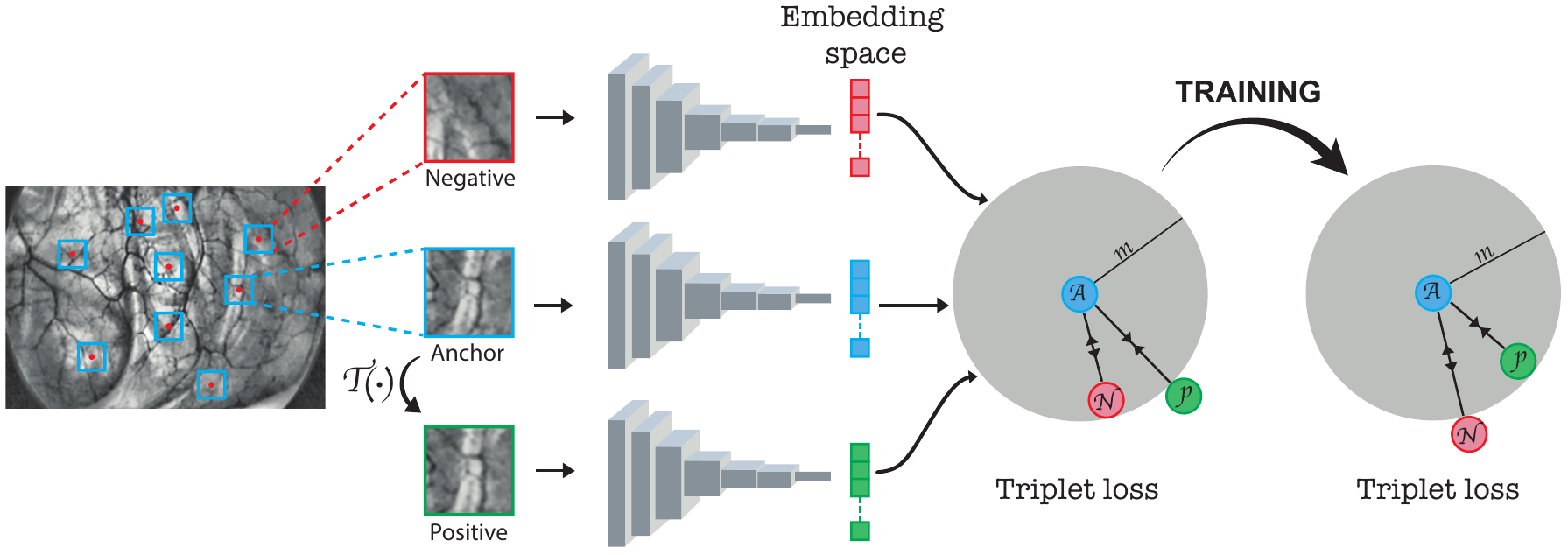}
\caption{Overview of the triplet loss training approach. Anchor, positive and negative patches are fed to the CNN  and trained so that the anchor-positive distance is minimized and the anchor-negative one is maximized in the feature space. As a result, a more discriminative embedding space is learned.}
\label{fig:training_phase}
\end{figure}
As depicted in \figureabvr \ref{fig:training_phase}, the training requires a triplet input image patches. For sake of simplicity, we note them $(P_{i} , P_{i}^{+},P_{i}^{-})$ where $P_{i}$ is the anchor patch, $P_{i}^{+}$ is the positive patch (has a similar appearance as  $P_{i}$) and $P_{i}^{-}$ is the negative patch.
An anchor patch is obtained by cropping a window around a given key-point. Its corresponding positive patch is generated by applying a homography transformation $T(.)$ to the anchor patch. $T(.)$ is a simulated transformation that often exists in endoscopic images. The negative patch could be any other patch selected from the same endoscopic image.

The method of selecting positive and negative patches to form triplets is critical, as random selection does not always produce good results. Various triplet selection approaches have been proposed in several studies \citep{hermans2017defense, cui2016finegrained, Yu18}. In our case, we use the same strategy proposed in recent HardNet method \citep{mishchuk2018working}. It aims at minimizing the distance between anchor and positive patches while maximizing the distance between the anchor and  the nearest negative patch. The loss equation is defined as follows:
\begin{equation}
L=\frac{1}{N_t} \sum\limits_{i=0}^{N_t} max\left(0, m+ d( f(P_{i}) - f(P_{i}^{+}))-min( d(f(P_{i})-f(P_{j_{min}}^{+})),d( f(P_{k_{min}})-f(P_{i}^{+}))\right)
\label{eq:hardnet_loss}
\end{equation}

\noindent where $d(f(P_{i}),f(P_{i}^{+}))=\sqrt{2-2f(P_{i})f(P_{i}^{+})}$, $P_{j_{min}}^{+}$ is the second nearest neighbor to $P_{i}$ after $P_{i}^{+}$, $P_{k_{min}}$ is the nearest non-matching anchor to $P_{i}^{+}$,  $m$ is a margin scalar, $j_{min}$ and $k_{min}$ are defined respectively in equations \ref{eq:j} and \ref{eq:k}.

\begin{equation}
j_{min}=\underset{j=1..N_t,j\neq i} {\mathrm{argmin}}  d(f(P_{i})- f(P_{j}^{+}))
\label{eq:j}
\end{equation}

\begin{equation}
k_{min}=\underset{k=1..N_t,k\neq i} {\mathrm{argmin}} d(f(P_{k})- f(P_{i}^{+}))
\label{eq:k}
\end{equation}

\section{Experiments and results}
\label{sec:expermients_results}

The evaluation of the proposed image key-points matching is performed in several stages to
study different aspects of the approach. The evaluation experiments are conducted
in three experimental sets:
\begin{itemize}
    \item Intrinsic evaluation of our method: we assess the robustness of our method to typical geometric transformation variations between frames such as view point, scale, and blurring changes. In addition, we back up our HardNet loss choice by evaluating different triplet loss variants.
    \item Comparison to state-of-the-art local feature descriptor methods: we consider both handcrafted and deep learning methods.
    \item Use-case of endoscopic image mosaicing: we qualitatively demonstrate the efficiency of our image key-points matching in the use case of endoscopic image mosaicing.
\end{itemize}
Before delving into the details of these experiments, we will first present our dataset, and go over our training settings.

\subsection{Database description}

\begin{figure}[h!]
\centering
\includegraphics[draft=\figuredraft,width=\textwidth]{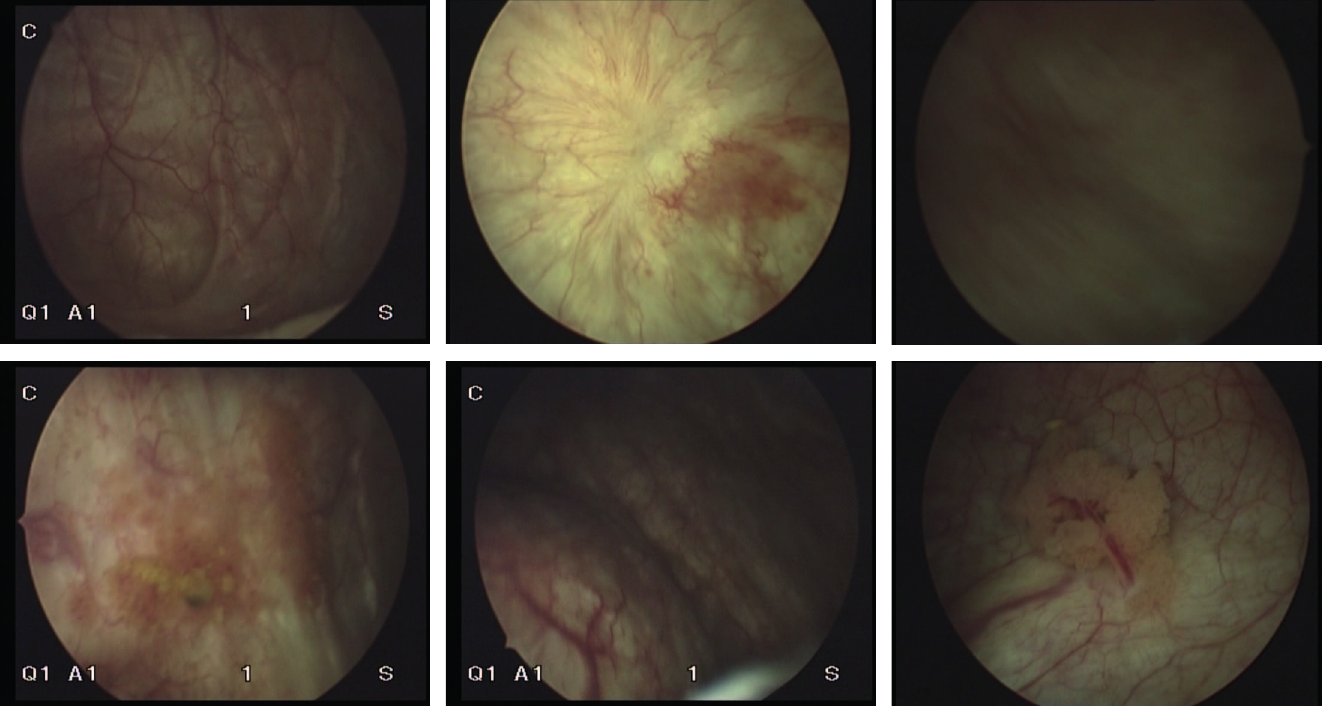}
\caption{Samples of clinical endoscopic images taken from different patients.}
\label{Fig:15}
\end{figure}

To generate our training database, five human bladder endoscopic videos from different patients acquired with the same endoscope have been used. Samples of such images are shown in \figureabvr \ref{Fig:15}. In all of the conducted experiments, we use 4 out of the 5 videos for training and keep the remaining video for validation in a cross-validation scheme.

We first convert raw data into gray-scale images and enhance contrast based on CLAHE method. Comes after key-points detection and anchors cropping.
As explained in the previous section, positive samples are obtained after applying any appropriate transformation to endoscopic frames. In our case, we applied rotations with small angles  ($\theta \in [5,10,15]$ degrees), scaling with small scale factor ($S_{f}=[0.9, 0.95,1.05,1.1, 1.15]$) and/or  small translations in both axes (\eg 8 pixels). In total, we created a training dataset composed of about 20k patch.

\subsection{Training settings}
The database was trained using Stochastic Gradient Descent with a batch size of 128 and an initial learning rate of 0.001 and a momentum set to 0.9. These are the final hyperparameters we considered after several runs, with the goal of improving convergence speed. The model converges in about 28 hours of training on a NVIDIA Titan V GPU with 12GB memory. The margin $m$ of equation \ref{eq:hardnet_loss} is experimentally fixed to 1. The  matching between $576\times720$  image pairs completes in less than 1 second. In our application context, the run-time is less important because panoramic image construction is typically done offline.

\subsection{Intrinsic evaluation of the proposed method}
\subsubsection{Robustness to image transformation variations}
To evaluate the robustness of the proposed local feature matching method against image transformation variations, we applied various combinations of geometric transformation to each image from the validation video in order to assess the robustness of our method against specific transformations (\ie viewpoint change, scale change, blurring).
Performances have been evaluated in terms of \textit{recall} with regards to \textit{precision} \citep{MikolajczykSchmid2005}. The number of correct and false matches are determined by the projection error PE instead of the overlap error used in \citep{MikolajczykSchmid2005}.
The projection error PE is defined as the Euclidean distance calculated between the matched key-points and the ground truth key-points (correct matches). In our experiments, we fixed experimentally the projection error to $PE=5$. We also report the obtained results for well-known handcrafted descriptors (\ie AKAZE, KAZE, SURF, SIFT, ORB and BRISK) to put our results into perspective. It is worth noting that since each original handcrafted descriptor is built on top of its own key-point detector, we decided to output the results for each of these detectors for a fair comparison. As an example, when applied to key-points detected with the code provided in the AKAZE implementation, we refer to our method as Proposed$_{AKAZE}$.

\subsubsection{Robustness to viewpoint changes}
\begin{figure}[ht!]
\centering
\includegraphics[draft=\figuredraft,width=1\textwidth]{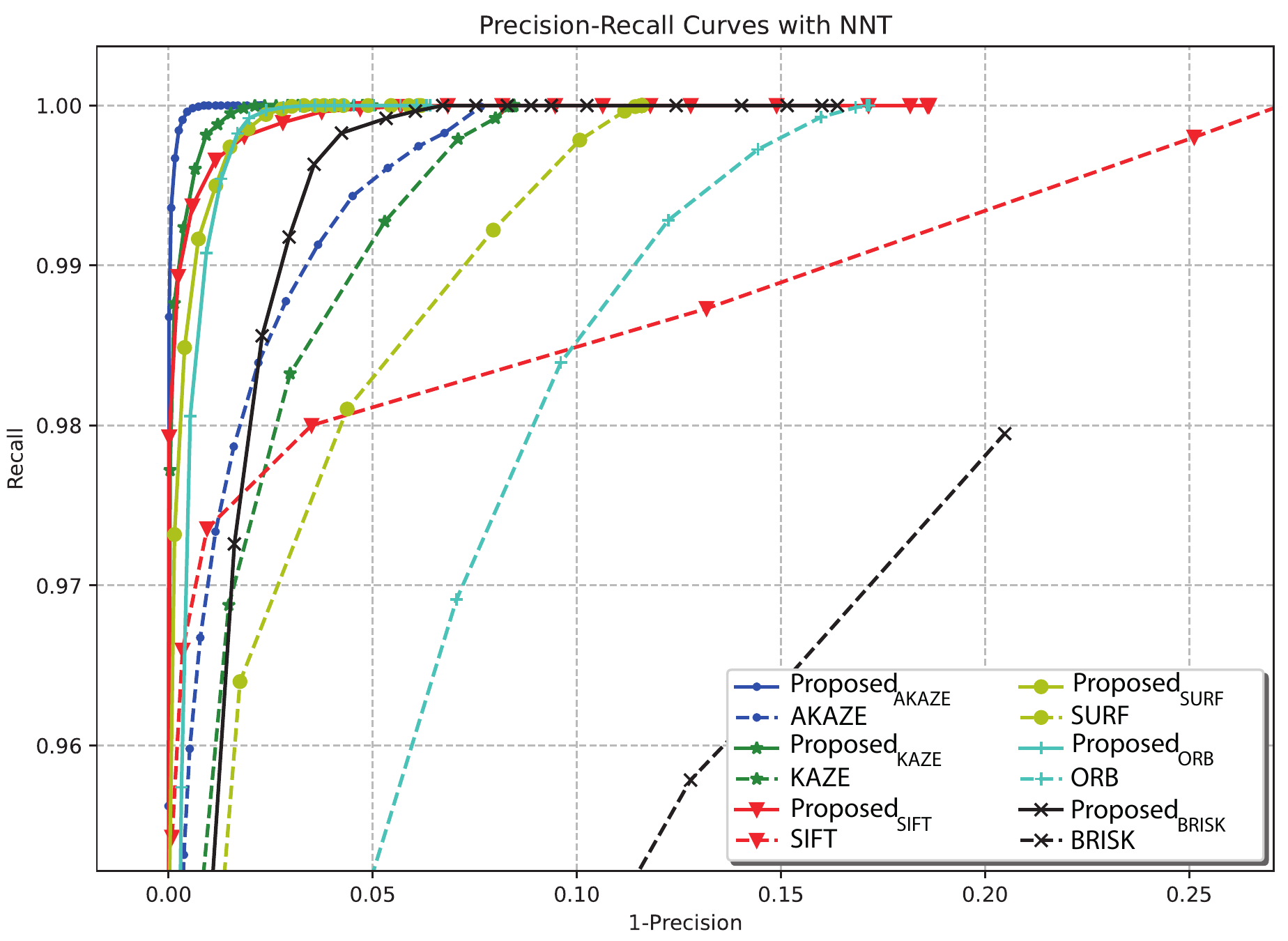}
\caption{Recall/precision curves computed for all descriptors when varying viewpoints. The proposed descriptor outperforms the handcrafted ones, reaching the maximal recall value for a precision rate of $99\%$ (when AKAZE is used as detector).}
\label{Fig:3}
\end{figure}

To simulate typical viewpoint changes in endoscopic videos, we estimate geometric transformations between consecutive frames in clinical data yielding in a bench of homography $3\times3$ matrices. This estimation is performed in a standard registration scheme based on key-points detection, matching, and RANSAC algorithm  \citep{MartinandRobert1981}. We applied a set of 10 randomly selected pre-computed transformations to each image from the validation endoscopic videos. Then, for each pair of images (input and transformed images), we detect key-points, extract local features and match descriptors according to the baseline to evaluate. The obtained recall and precision results are shown in \figureabvr \ref{Fig:3}.

\figureabvr \ref{Fig:3} shows that the proposed descriptor outperforms the handcrafted ones. Indeed, the maximal recall value is reached for a precision value of about $99\%$ for Proposed$_{AKAZE}$ setting. We notice also that AKAZE is outperforming the other handcrafted descriptors.

\subsubsection{Robustness against scale changes}

\begin{figure}[h!]
\centering
\includegraphics[draft=\figuredraft,width=0.95\textwidth]{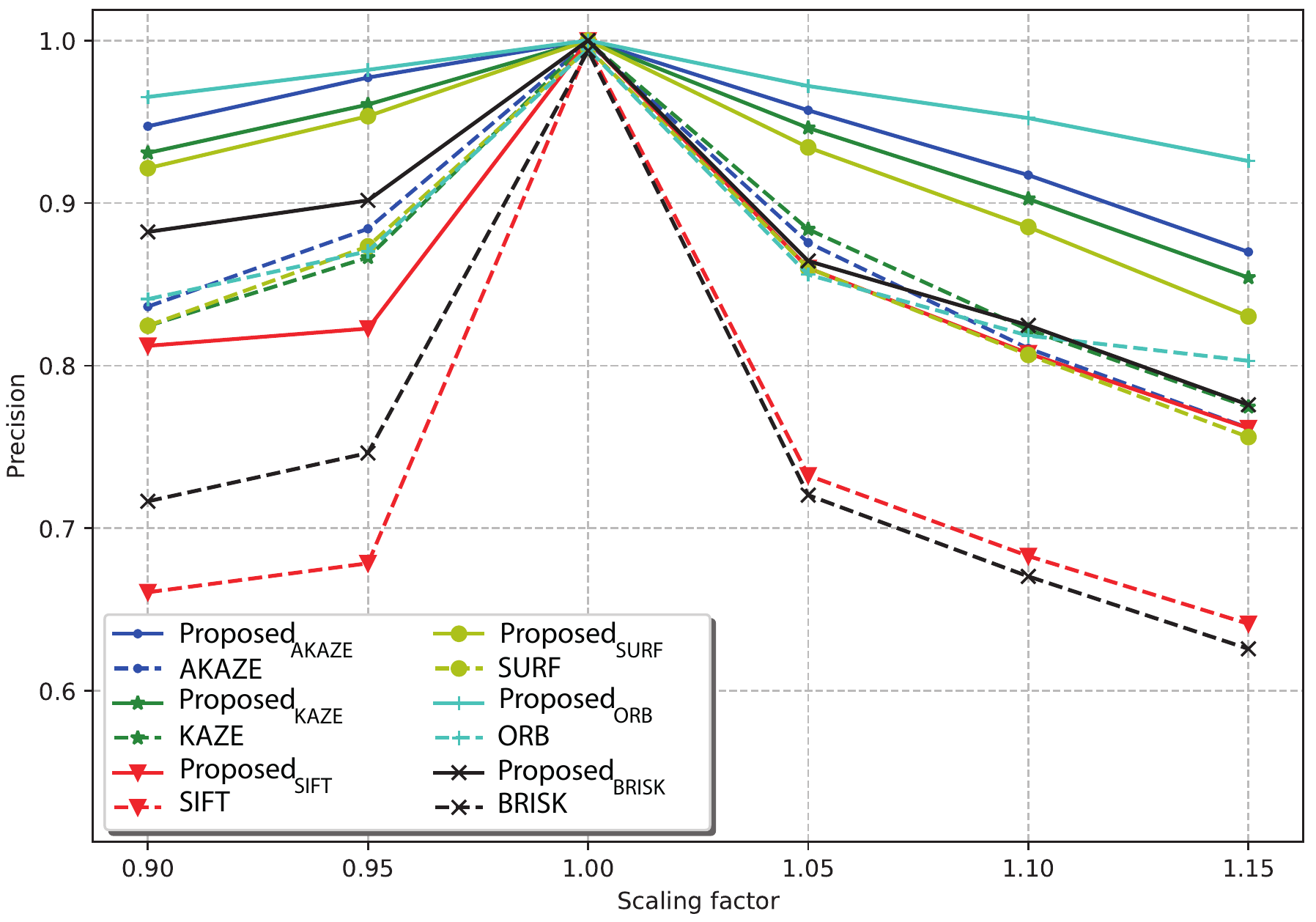}
\caption{Precision rates computed for different scale factors. The proposed descriptor outperforms the handcrafted ones, reaching higher precision rates, showing thus less sensitivity to scaling changes.}
\label{Fig:4}
\end{figure}

Considering that only small variations occur generally between two consecutive frames in endoscopic videos, we do not need to evaluate large scale changes. We evaluated the performance of the different descriptors only against small pre-defined scale factors $\{0.9,0.95,1,1.05,1.1,1.15\}$. 
The result relative to the sensitivity of the proposed descriptor to scale changes is depicted in \figureabvr \ref{Fig:4}

\figureabvr \ref{Fig:4} highlights that handcrafted descriptors are more sensitive to scaling factor changes than the proposed descriptor. Especially, for Proposed$_{ORB}$, Proposed$_{AKAZE}$, Proposed$_{KAZE}$, and Proposed$_{SURF}$ configurations, the precision of the proposed descriptor remains high (almost more than $90\%$) for scaling factors in the range of $[0.9,1.1]$. 

\subsubsection{Robustness against blurring}

Endoscopic videos are often captured with blur due to several factors such as small hand vibration, bad choice of lens focus, fast movement of endoscopic camera with a low frame rates, \etc Therefore, we have to quantitatively measure the descriptors efficiency regarding the blur artifact. Thus, we blurred the input video frames by applying various convolution kernels $(3\times3, 5\times5,10\times10,15\times15)$.

\begin{figure}[h!]
\centering
\includegraphics[draft=\figuredraft,width=12cm]{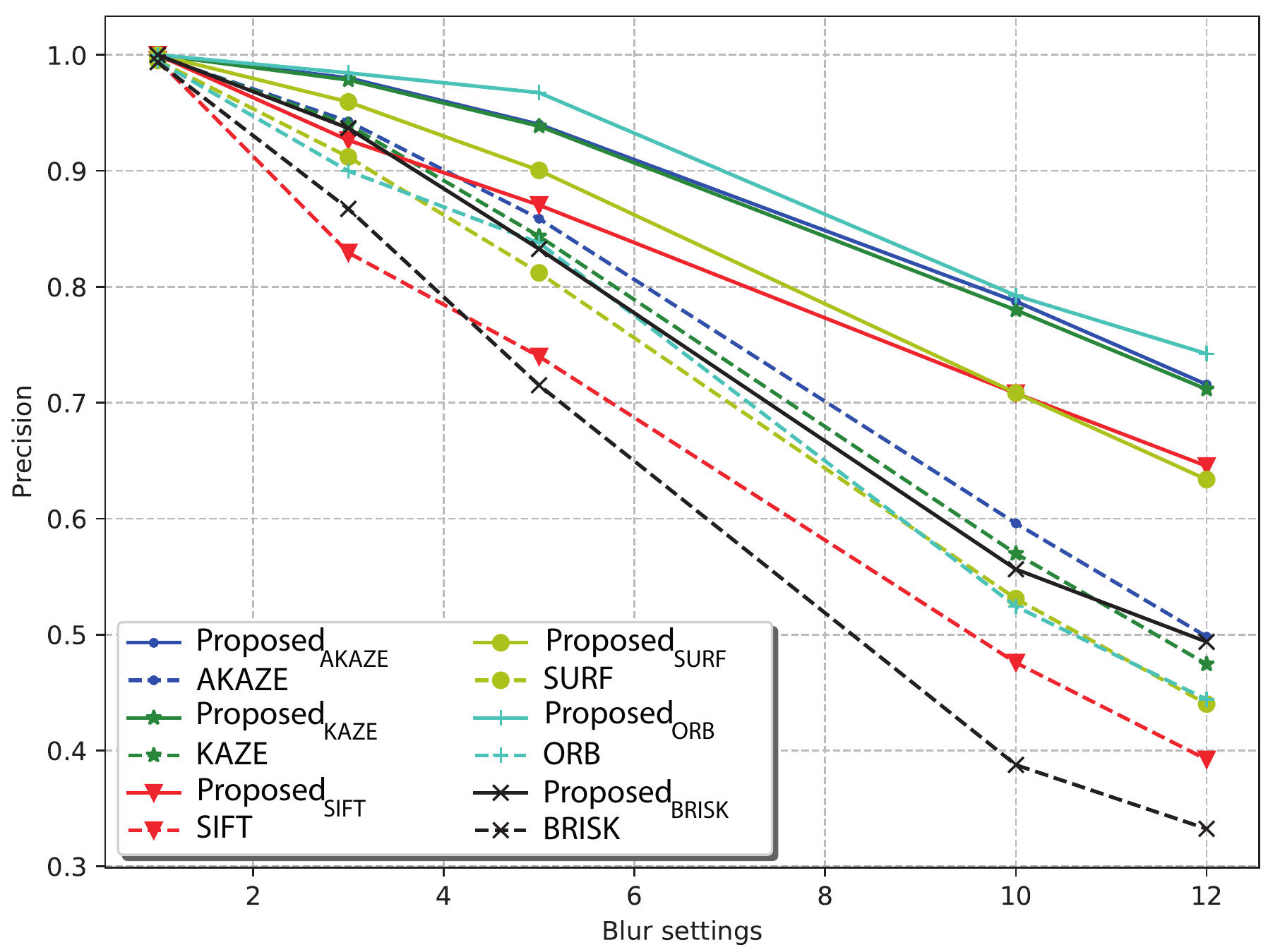}
\caption{Precision rates computed with different blur created with different kernel sizes. The proposed descriptor is the least sensitive to blur. }
\label{Fig:5}
\end{figure}

\figureabvr \ref{Fig:5} illustrates quantitatively how blur affects matching precision. Compared to handcrafted descriptors, the proposed descriptor is the least sensitive to blur.

\subsubsection{Comparison of triplet loss variants}
In here, we consider different triplet Loss functions: the HardNet Loss \citep{mishchuk2018working}, the standard Triplet Loss with a fixed margin \citep{Schroff_2015}, and the adaptive margin triplet Loss \citep{WANG2018241}. The HardNet and the triplet loss functions are tested with a same margin $m=1$. 
Similarly to \citep{mishchuk2018working}, we evaluate the different losses in terms of precision which is defined as the ratio of correct matches over the total number of matches and matching score  which is the ratio of correct matches over the total number of detected key points. 
The obtained comparison results  are illustrated in \tableabvr \ref{table:2} showing the outperformance of HardNet loss which explains our decision to select it for our CNN model training.

\begin{table}[]
\caption{Comparison of three different loss functions: HardNet Loss, Triplet Loss and the adaptive margin triplet Loss in terms of precision and score matching in $\%$. HardNet loss shows better performance.}
\label{table:2}
\centering
\begin{tabular}{lcc}
\hline
& \textit{Precision}     & \textit{Matching Score}             \\ \hline
HardNet Loss  & {\textbf{99.01}} & \textbf{80.36}    \\
Triplet Loss  & {95.80}          & 79.13             \\
Adaptative margin triplet Loss & {92.19}          & 80.34             \\ \hline
\end{tabular}
\end{table}

\subsection{Comparison to state-of-the-art local feature descriptor methods}

There is no publicly available benchmark for evaluating endoscopic image matching at the moment. We decided in these experiments to annotate the frames of one validation video to cover more realistic transformations between consecutive frames. The annotation is carried out as follows. We first detect key-points in all the video frames, and then we manually annotate the correspondence between key-points with a customized tool yielding over 1000 annotated image pairs.

\subsubsection{Comparison to state-of-the-art Handcrafted descriptors}
\begin{figure}[h!]
\centering
\includegraphics[draft=\figuredraft,scale=1]{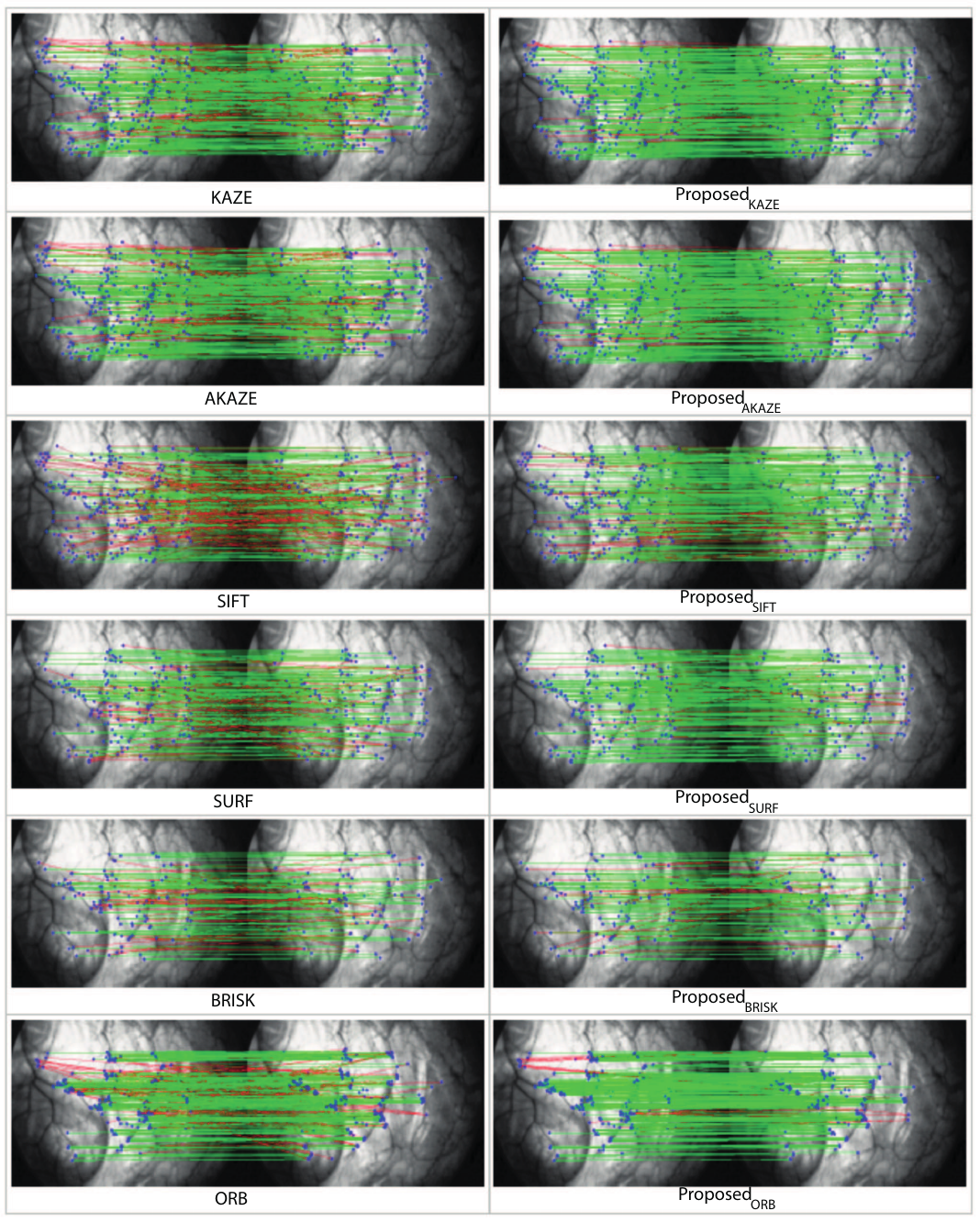}
\caption{Qualitative evaluation of the proposed descriptor. Green  and red lines refer respectively to correct and wrong matches. The best matching quality (number of green lines) is reached with the proposed descriptor.}
\label{Fig:6}
\end{figure}

These experiments are being carried out to consolidate the results previously obtained in the intrinsic validation of our method, demonstrating its superiority over all well-known handcrafted descriptors.
In \figureabvr \ref{Fig:6}, we depict the matching results between two endoscopic frames. Green  and red lines refer respectively to correct and wrong matches.  Compared to all the used descriptors, we can observe that a larger number of correct matches (green lines) is achieved with our approach.

\begin{figure}[h!]
\centering
\includegraphics[draft=\figuredraft,scale=0.65]{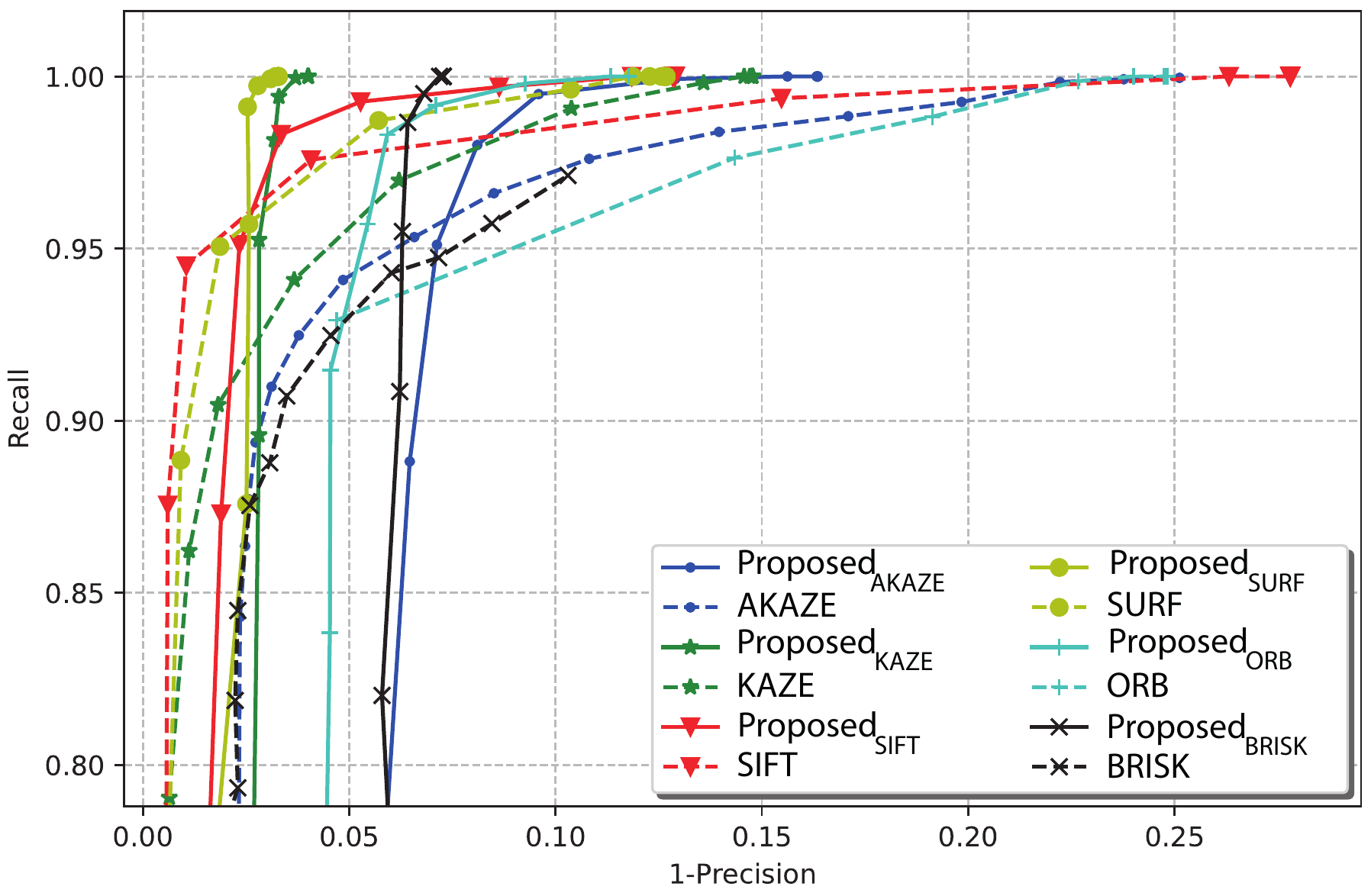}
\caption{Quantitative evaluation of the proposed descriptor on real clinical dataset.  The maximal recall rate  could be achieved for a precision rate of almost $98\%$ when our descriptor is used with SURF or KAZE detectors (\ie Proposed$_{SURF}$ and Proposed$_{KAZE}$, respectively.}
\label{Fig:7}
\end{figure}

For quantitative evaluation, we report the obtained matching performances in terms of recall and precision in \figureabvr \ref{Fig:7}.

From \figureabvr \ref{Fig:7}, we can observe that the maximal recall rate could be achieved for a precision rate of almost $98\%$ when our descriptor is paired with SURF or KAZE. However, when handcrafted descriptors are used, the best recall value could be obtained for a maximal precision rate of $88\%$ (in the case of SURF). A trade-off between precision and recall should be undertaken to choose the best descriptor.

\subsubsection{Comparison to deep learning based \rev{supervised} method\rev{s}}
As previously stated, no benchmarking for endoscopic image key-points matching exists in the literature. Nonetheless, to put our results into perspective, we compare the performance of our descriptor to \rev{the most recent state-of-the-art methods for key-point matching based on supervised learning:
SosNet \citep{sosnet2019cvpr}, SuperGlue \citep{sarlin2020superglue}, HyNet \citep{hynet2020}, and CNDesc \citep{CNDesc9761761}.} We \rev{used} 800 image pairs for training and \rev{left} the remaining 200 images for testing.

\rev{Both SuperGlue and CNDesc originally require first to extract a set of feature key-points from the training and testing images. We followed the same procedure to train SuperGlue and CNDesc on our dataset. In our experiments, feature key-points were detected using SIFT method as we did not observe any significant performance differences when choosing different key-point detectors. On the other hand, training the HyNet and SosNet models as well as our proposed model requires a set of extracted patches based on the localization of image key-points. To this end, the training patch dataset is extracted around the SIFT image key-points that have already been detected. The training for all five competing methods is therefore based on the same key-points.}

Comparative results in terms of precision and matching score metrics are summarized in \tableabvr \ref{table:1}. \rev{Despite being the most recent work, CNDesc shows the lowest results and seems to be the least suitable to endoscopic videos. The other methods achieved competitively high performances. However, the fundamental difference between our method and the other competing ones is still the training mode and our ability to adapt automatically to the large texture variability in endoscopic videos. Indeed, we remember that in our case the training triplets are automatically generated which means that a fine-tuning step could be carried out whenever needed on new endoscopic videos without any annotation.}

\begin{table}[ht!]
\caption{Comparison between our method \rev{and the most recent state-of-the-art methods for key-point matching based on supervised learning} in terms of precision and matching score reported in (\%). \rev{We notice that all the obtained results are very comparable.}}
\label{table:1}
\centering
\begin{tabular}{lcc}

\hline
& \textit{Precision}     & \textit{Matching Score}  \\ \hline
\rev{SosNet \citep{sosnet2019cvpr}}& \rev{99.95}            &\rev{91.60 }           \\
\rev{SuperGlue \citep{sarlin2020superglue} }   &\rev{ 99.17     }        & \rev{92.81 }           \\
\rev{HyNet \citep{hynet2020} }  & \rev{99.96}            & \rev{90.29}            \\

\rev{CNDesc \citep{CNDesc9761761}}   & \rev{98.37}            &\rev{90.51}           \\
Proposed     & \rev{99.89}    & \rev{92.56} \\ \hline
\end{tabular}
\end{table}

\subsection{Use-case of endoscopic image mosaicing}
\begin{figure}[h!]
\centering
\includegraphics[draft=\figuredraft,width=10cm]{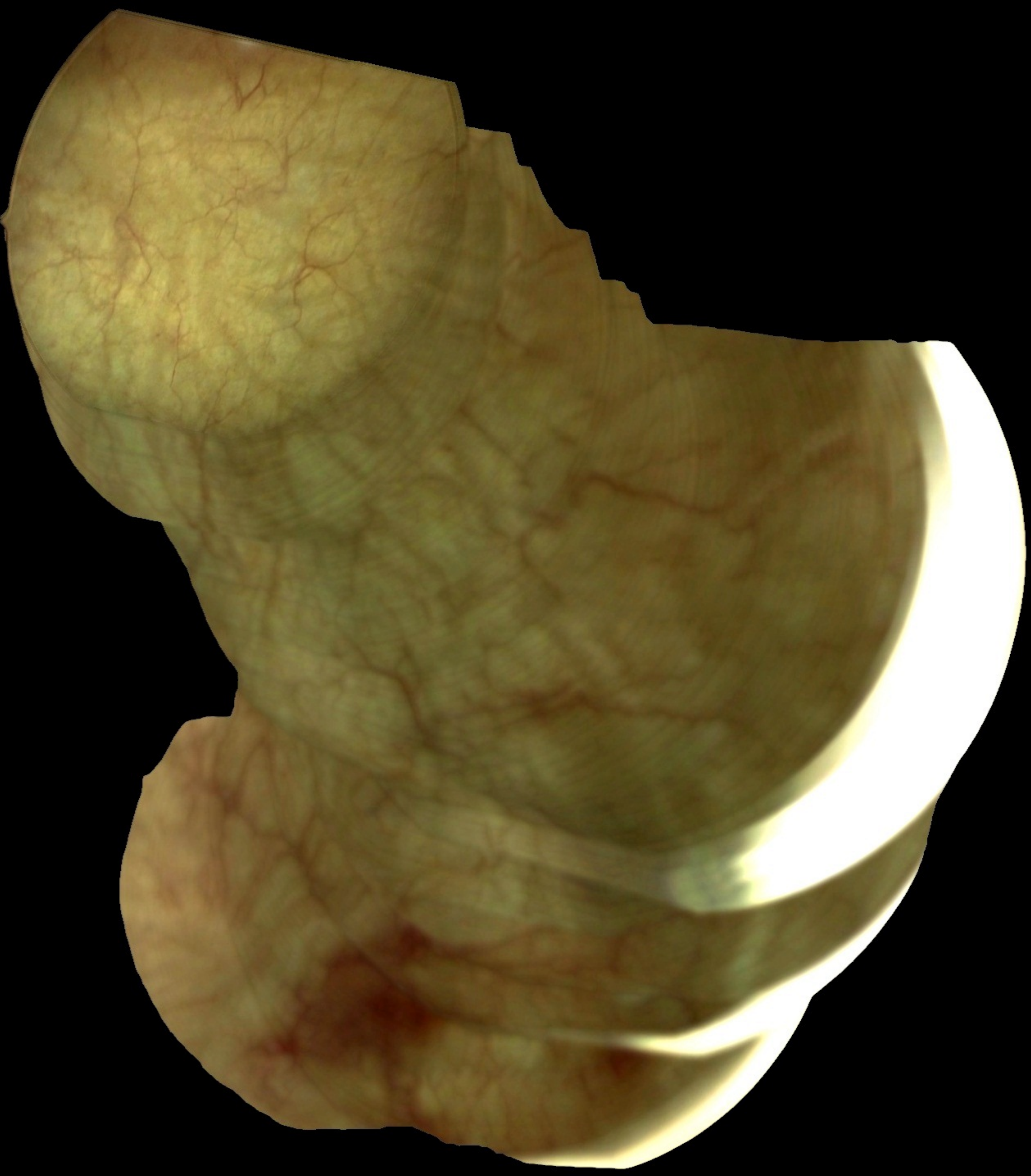}
\caption{A $1258\times1436$ panoramic image constructed from a 400 consecutive frames from human bladder endoscopic video. The obtained mosaic image is coherent and we can observe clearly the texture continuity which is a good indicator proving the precision of the images alignment.}
\label{Fig:9}
\end{figure}

To validate the effectiveness of the proposed descriptor in generating panoramic images, we feed the obtained matching results to a mosaicing system based on RANSAC algorithm \citep{MartinandRobert1981}. We construct a panoramic image with a set of 400 consecutive frames from bladder endoscopic video, as shown in \figureabvr \ref{Fig:9}.

The obtained mosaic image is coherent and we can observe clearly the texture continuity which is a good indicator proving the precision of the images alignment. This experiment illustrates that the proposed descriptor provide reliable and robust matching features  that can be utilized to construct panoramic images for endoscopic videos. However, a strong blur effect or large scale change can perturb the registration process and cause discontinuities in the resulting mosaic image. To overcome this problem, we can extend the mosaic image system with an artifact detection algorithm.

\section{Conclusions and perspectives}
\label{sec:conclusions}

Despite the success of deep learning approaches in a variety of computer vision tasks, a lack of labeled data remains a major barrier to the use of neural networks in medical applications. To address this issue, we proposed a self-supervised approach for endoscopic image matching in this paper, which is based on the automatic generation of a pseudo labeled data-set. As a result, our method allows us to train a local descriptor network using only endoscopic images, with no need for labeled data or manual annotation. The proposed self-supervised approach was evaluated and compared to different handcrafted image feature descriptors and also to \rev{recent deep learning based supervised methods: SosNet, SuperGlue, HyNet and CNDesc.}

The experimental results proved the robustness of our descriptor against viewpoint, scaling factor, and blurring changes. \rev{Moreover,  compared to the supervised state-of-the-art deep learning based methods, our approach achieves
competitive performance in terms of precision and matching score while using unlabeled patches in a self-supervised training mode.} For future works, we would like to investigate further our approach on endoscopic videos of other organs such as small intestine, large intestine and stomach. In addition, we can use different endoscopy system such as capsule endoscopy or blue laser endoscopy system.

\bibliography{submission_R2}

\end{document}